10th International Conference on Information Technology and Quantitative Management

# Review of medical data analysis based on spiking neural networks

Li Xiaoxue[a]，Zhang Xiaofan[a]， Yi Xin[a]，Liu Dan[a]，Wang He[a]，Zhang Bowen[a]，

Zhang Bohan[a]，Zhao Di[c,d]，Wang Liqun[a,b,*]

[a]College of science，China University of Petroleum，Beijing 102249，China

[b]Beijing Key Laboratory of Optical Detection Technology for Oil and Gas, China University of Petroleum, Beijing 102249 China

[c]Institute of Computing Technology，Chinese Academy of Sciences，Beijing 100190，China

[d]University of Chinese Academy of Sciences, Beijing 100049, China

**Abstract**

Medical data mainly includes various types of biomedical signals and medical images, which can be used by professional doctors to make judgments on patients' health conditions. However, the interpretation of medical data requires a lot of human cost and there may be misjudgments, so many scholars use neural networks and deep learning to classify and study medical data, which can improve the efficiency and accuracy of doctors and detect diseases early for early diagnosis, etc. Therefore, it has a wide range of application prospects. However, traditional neural networks have disadvantages such as high energy consumption and high latency (slow computation speed). This paper presents recent research on signal classification and disease diagnosis based on a third-generation neural network, the spiking neuron network, using medical data including EEG signals, ECG signals, EMG signals and MRI images. The advantages and disadvantages of pulsed neural networks compared with traditional networks are summarized and its development orientation in the future is prospected.

## 0 Preface

With the development of science and technology and the improvement of living standards, human pursuit of health is also increasing. Currently, the technology of analyzing physical conditions using biomedical data such as electroencephalogram (EEG), electrocardiogram (ECG), Electromyographic signal (EMG) and magnetic resonance image (MRI) is relatively mature. However, these interpretation tasks are time-consuming and tedious, which require professional medical personnel. In order to improve efficiency and accuracy to alleviate doctors' workload, it is inevitable to apply computer-aided diagnosis (CAD) to medical health.

*Corresponding author.
E-mail address: wliqunhmily@cup.edu.cn.

In recent years, there have been many studies using machine learning for biomedical image and electrical signal analysis [1, 2]. However, although traditional artificial neural networks are inspired by biological neurons, they do not have biological interpretability and require a large amount of computation and energy consumption, which is not conducive to real-time and rapid analysis of medical data. With the development of neural networks, the third-generation neural network――Spiking Neural Network (SNN) has emerged. Although its accuracy is relatively low and there are difficulties in training, SNN has the advantages of lower energy consumption, faster speed, and better applicability to spatiotemporal data due to its more biologically interpretable network structure and training rules. Therefore, using spiking neural networks for research on medical data is of great significance.

This article introduces the basic structure and common algorithms and models of spiking neural networks, and reviews the latest research progress and applications of spiking neural networks in medical data, including EEG, EMG, ECG, MRI, etc. Finally, these analysis methods are summarized and future research trends and possible development directions are predicted.

**1 Theoretical basis of spiking neural network**

*1.1. The structure of spiking neural network*

The biological neurons in the brain are mainly composed of three parts: dendrites, somatic cells, and axons. Dendrites receive input signals, and transmit signals from other neurons to the somatic cells. The somatic cell is the central processing unit responsible for information processing. The output from the somatic cell is in the form of pulses that are released via axons. When a post-synaptic neuron receives these pulses, the membrane voltage changes. If the voltage surpasses the threshold, the post-synaptic neuron will release a pulse [3]. Hodgkin and Huxley [4] conducted experiments on the giant axons of squid, mainly studying the relationship between potassium, sodium, chloride ions, and membrane voltage. They abstracted biological cells as neuron models, known as the Hodgkin-Huxley (HH) model, which has a lot of biological details and high computational cost. Afterward, many simplified neuron models were proposed, such as the Leaky Integrate-and-Fire (LIF) model [5] and the Izhikevich neuron model [6].

Spiking neurons are a simplified version of biological neurons. They are connected through synapses and transmit information through pulse signals. The pulses can be either excitatory or inhibitory, and the synaptic weight can be automatically adjusted through network learning [3].

*1.2. Introduction to common algorithms and models of SNN*

*1.2.1. Spike Timing Dependent Plasticity（STDP）*

Spike Timing Dependent Plasticity (STDP) is a unique learning rule of pulse neural networks. It is in accordance with the cognitive and learning rules of human brain neurons. STDP adjusts the synaptic weight based on the time difference between the pre-synaptic and post-synaptic neurons firing spikes. When the pre-synaptic neuron fires spikes shortly before the post-synaptic neuron, it causes long-term potentiation (LTP), which corresponds to an increase in the synaptic weight. Conversely, if the pre-synaptic neuron fires spikes after the post-synaptic neuron, it causes long-term depression (LTD), leading to a decrease in the synaptic weight, and the shorter the interval time, the greater the amount of change [7].

*1.2.2. NeuCube model based on SNN architecture*

Kasabov (2014) proposed the NeuCube model, which is mainly used to process spatiotemporal data, such as EEG, fMRI, etc. NeuCube is a 3D-Spiking Neural Network (3D-SNN) model based on brain structure, which is divided into three main modules:

1. Encoding module: encodes spatiotemporal data into pulse sequences and maps them to the next module.
2. 3D reservoir module (SNN reservoir, SNNr): SNNr is a 3D spiking neural network constructed based on brain structure, with neurons initially connected according to the small-world principle [8] and trained using the unsupervised STDP rules.
3. Classification module: generally supervised learning is performed by evolving SNN (eSNN) classifier [9].

Among them, the SNNr module constructs a 3D model similar to the brain structure, which not only improves the classification accuracy of spatio-temporal data, but also shows the activity trajectory among neurons, reflecting the connectivity of different functional areas of the brain, making the neural network no longer a black box, but a good interpretive one, helping humans better understand the functions of the brain.

## 2 Electroencephalogram (EEG) analysis

The brain is one of the most important parts of the human body, and electroencephalogram (EEG) is an electrical signal generated by the daily activities of the brain, which can be used for classification tasks, disease research and prediction. EEG signals can be obtained by placing electrodes on the scalp for data acquisition. At present, a large number of different types of EEG data sets have been collected for research, such as DEAP data set for emotion recognition [10], sports imagination dataset BCI Competition IV [11], sleep data set Sleep-EDF [12] and so on.

### 2.1. Task of classification

#### 2.1.1. Emotional recognition

Emotion is the psychological response of human beings to external stimuli, and bad emotions are not conducive to people's physical and mental health, so it is necessary to identify emotions. Correct emotional representation is the premise of emotional recognition. Emotional representation methods are mainly divided into two categories: (1) Discrete models. The model regards complex emotions as a combination of basic emotions. (2) Continuous model. The continuous model maps emotions into two-dimensional or three-dimensional models, and uses continuous variables to represent emotions.

Tan et al. [13] used electroencephalography (EEG) to achieve short-term emotion recognition based on the NeuCube framework. The segmentation of EEG is realized by using facial feature points, which saves the steps of artificial EEG segmentation. The experimental results on DEAP and MAHNOB-HCI data sets [14] show that the classification accuracy of arousal degree is 78.97% and 79.39%, and the classification accuracy of valence is 67.76% and 72.12% respectively. The visualization of this model is helpful for us to understand the brain working mechanism corresponding to different emotions.

He et al. [15] also used the DEAP dataset for model training and classification based on the NeuCube framework. Four of the 32 electrode channels in the DEAP dataset - FP1, FP2, F3, F4 - were selected to classify the valence into positive and negative categories. Step Forward (SF) coding is adopted for coding, and the parameters are optimized by grid algorithm. By using K- nearest neighbor algorithm, 32*40 sample data from four electrode channels of 32 testers were verified. The average accuracy rate was 68.91%, and the highest classification accuracy rate reached 82.50%.

Yan et al. [16] used a migration learning approach to train convolutional neural network (CNN) networks first and then pass the weights of CNNs to equivalent SNNs to achieve conversion of CNNs to SNNs to finally achieve classification of EEG signals. The final performance of the experimental results is that the accuracy of 82.75% and 84.22% are achieved on the validity dimension and arousal dimension of the DEAP dataset respectively, while the energy consumption is only 13.8% of that of the CNNs.

#### 2.1.2. Cognitive classification

Cognitive process refers to imagining motor movements in the brain without physical movements. Kasabov et al. [17] used the NeuCube model to classify five types of cognition collected from seven subjects, including relaxation, imaginary letter writing, multiplication, counting, and graph rotation. This study proves that it is feasible to use NeuCube model to model and analyze EEG data, which can provide a new method for analyzing other spatiotemporal data.

Based on the framework of NeuCube, Doborjeh et al. [18] classified and analyzed the EEG data collected by three groups of addicts in Go/Nogo cognitive tasks, namely, untreated opioid addicts, opioid addicts receiving methadone treatment (MMT) and healthy controls, and visualized the EEG by using the model, revealing that MMT has a potential positive effect on brain function, which is conducive to the recovery of brain function.

*2.1.3. Motor imagery*

Motor imaginery EEG refers to the EEG signals when people imagine some sports. The connection between the brain and external devices is established through EEG signals, which can help people by controlling the prosthetic devices and other auxiliary devices is realized.

Virgilio et al. [19] used the BCI Competition IV dataset to perform a two-stage binary classification for each of the five motor imagery (MI) tasks - left hand MI, right hand MI, foot MI, tongue MI and rest - by comparing traditional neural network models with impulse neural network, it was demonstrated that the best classification results were obtained using SNNs containing temporal feature data. The first of these stages considered only static features of the data and compared the SNN model with the traditional MLP network. The classification results showed that the SNN outperformed the MLP model in seven of the ten classification scenarios, with an accuracy of 80% in most cases. The second stage compares the performance of the SNN model with different inputs. The experimental results showed that using the original EEG signal gave better classification results than feature extraction, that data containing temporal features had improved classification results compared to static data results, and that the average accuracy of the SNN reached 83.16%. It proved that SNNs have the advantage of spatio-temporal information analysis and can achieve better results than traditional models using a smaller number of neurons.

Kumarasinghe et al. [20] proposed a brain-inspired impulse neural network model (BI-SNN) based on the NeuCube framework and eSPANNet [21] to achieve muscle movement prediction of EEG signals. This was done by replacing the eSNN in the NeuCube framework for supervised learning with eSPANNet (evolved spiking pattern association neural network), which was supervised trained using the EMG signals and kinematic signals as the expected output. Results show that BI-SNNs have lower latency and better accuracy, providing a promising direction for brain-computer interfaces.

*2.2. Epilepsy*

Epilepsy is a chronic neurological disease. According to the World Health Organization in 2022, there are approximately 50 million epilepsy patients worldwide. The symptom of epilepsy is muscle twitching, and repeated and sudden attacks can be life-threatening. Therefore, timely early warning of epilepsy is necessary.

Deep neural networks (DNNs) are already highly accurate in predicting seizures [22], but the application of DNN-based methods in healthcare is limited by the drawbacks of high energy consumption and high latency of DNN operations. Based on the characteristic of low latency and low energy consumption of pulsed neural networks, Zarrin et al. [23] implemented classification of EEG signals through pulsed neural networks to achieve epilepsy prediction. The gradient substitution method - the sigmoid function - was used to achieve back propagation of the network. Analysis of the iEEG data provided by the Kaggle competition containing three types of signals regarding seizures, preictal and interictal, resulted in a test set accuracy of 97.6%, with a corresponding DNN accuracy of 95% and a convolutional DNN accuracy of 99.6%, but the latter two require huge computational effort and energy consumption and are not suitable for wearable devices. Therefore, although some accuracy is sacrificed, pulsed neural networks are more suitable for wearable devices for data processing, thus enabling monitoring and early warning of diseases in daily life.

Monitoring is essential for the treatment of epilepsy and high frequency oscillations (HFO) between seizures recorded in the EEG are potential biomarkers of epileptogenesis, seizure propensity, disease severity and response to treatment [24, 25] and can be used to analyse patients' seizures by detecting HFO. Burelo et al. [26] constructed two parallel SNN networks - a core SNN and an artifact detection SNN - to analyse 20 cases of EEG data from 11 children with epilepsy. The core SNN receives EEG in the 80-250 Hz band and the network is divided into two main layers, where the input layer consists of two neurons that generate positive and negative pulses, respectively. The output of the core SNN is an event of interest (EOI). The artifact detection SNN input signal is in the 500-900 Hz band. The purpose of the artifact detection SNN is to capture HFO-like oscillations, and if a signal is detected at this stage, the previous EOI output from the core SNN will not be classified as an HFO, and if no artifact is monitored, the previous EOI will be classified as an HFO. The results of the study show that the occurrence of detected HFOs correlates with epileptic activity with an accuracy of 80%.

*2.3 Sleep staging*

Sleep is an essential life activity for humans and there are currently many people worldwide with varying sleep problems, making the detection of sleep of great practical significance. Polysomnography (PSG) is the recording of several physiological parameters such as EEG, electrooculography and electromyography during sleep in a sleep laboratory, using computerised equipment. Polysomnography (PSG) can be used for sleep disease diagnosis and sleep medical research. Sleep staging can basically be divided into three stages [27]: wakefulness (WA), non-rapid eye movement (NREM) sleep and rapid eye movement (REM) sleep. The non-rapid eye movement (NREM) sleep stages can be subdivided into N1, N2 and N3, where N1 and N2 are light sleep stages and N3 is known as deep sleep stage or slow wave sleep (SWS). Different brain wave types occur in different sleep stages and doctors can classify the sleep stage according to the brain wave type, but this traditional expert labelling method is very inefficient and error-prone, so many people are now trying to use machine learning for automatic sleep staging. Traditional machine learning usually classifies by feature extraction [28], which tends to ignore spatio-temporal features, whereas impulse neural networks are more suitable for processing spatio-temporal data and are more suitable for automatic sleep staging.

Zhao et al. [29] performed sleep staging on PSG data based on SNN. The specific process was to first convert the PSG data into impulse data using the BSA algorithm, and then split the classification task into two stages: the first stage used a triple classification SNN to classify the samples into category A (WA), category B (N1&N2) and category C (SWS&REM), and then removed the samples from category WA for the second stage of quadruple classification to classify category B (N1&N2) and category C (SWS&REM) for subdivision. The experimental results show that the accuracy for the N2 and REM classifications are higher and the accuracy for the WA and SWS stages classifications are lower, with a final accuracy of 67.18%.

Budhraja et al. [30] implemented sleep staging of EEG data using NeuCube. The NeuCube framework was generally divided into three parts: pulse encoding, unsupervised training STDP, and supervised training DeSNN [31]. Three features of the NeuCube framework were extracted on Spinnaker: the number of pulses in the input pulse sequence (SC), the number of spikes that appear on each neuron during DeSNN (SPN), and the final weight in the DeSNN (FW), using six classifiers for each classification. Results showed that the gradient-enhanced decision tree GBDT performed best on the combined SPN + FW features, with an accuracy of roughly 80% for the sleep staging results, except during the N1 stage.

**3 The Analysis of Electrocardiogram (ECG)**

Arrhythmias are intermittent and may be difficult to detect by routine inspection, so it is necessary to conduct long-term monitoring of heart rate. Spiking neural networks have the characteristics of low energy consumption and low delay, and neuromorphic chips can be used to realize real-time monitoring of wearable devices.

Clinically relevant heartbeat features are extracted from heart rate (also known as RR interval). The characteristics of RR interphase can well distinguish the type of distraction jump. Balaji et al. [32] realized ECG classification by converting CNNs into SNNs. The accuracy of CNNs heartbeat classification is 85.92%, while SNNs losses accuracy of 0.6% to 1%. Compared with CNNs, although SNNs has a little loss in accuracy, the amount of neuron activation and synaptic triggering events are greatly reduced under spiking neural networks, and a considerable part of neurons are in a silent state, so the energy consumption of SNN is lower. Yan et al. [33] also realized the classification of ECG data in MIT-BIH database by converting CNNs to SNNs. The results show that in the binary classification, the average accuracy of CNNs is 81%, and the average accuracy drops to 79% after converting into SNNs, but the energy consumption of SNNs is only 0.7% of that of CNNs. In the f four- class classification, the accuracy of SNNs is comparable to CNNs, but its power consumption is only 0.9% of CNN. Therefore, under the condition that the accuracy rate remains basically unchanged, SNNs has lower power consumption than CNNs, which is more conducive to data processing of wearable devices.

Amirshahi et al. [34] used the MIT-BIH arrhythmia database to train STDP and reward-modulated STDP (R-STDP) network models to realize low-power, real-time ECG signal classification for wearable devices. There are three main parts: spiking coding, pattern extraction and classification. In this model, unsupervised STDP is used to extract patterns in the signal, assisted by Gaussian layer and suppression layer, where Gaussian layer is used to adjust the number of spikings, and suppression layer helps STDP to capture different patterns. Supervised R-STDP was used to classify the extracted patterns, and the predicted classification result was the category corresponding to the neuron with high firing frequency. If the prediction was correct, the synaptic weight of the neuron was increased, otherwise it was reduced. Finally, the energy consumption is reduced, and the accuracy is 97.9%, as high as the traditional neural network.

Das et al. [35] implemented heart rate estimates of internal clinical data and three publicly available datasets based on unsupervised learning liquid state machine (LSM). The three publicly available datasets are MIT-BIH Supareventricular arrhythmia database, MIT-BIH Long-Term Database (ltdb) and long-term ST database (ltstdb). Different types of subjects are selected from different datasets, and then the model is trained in three steps: spiking coding, spiking neural network and heart rate estimation. It is first time encoded, and then the spiking sequence is imported into a neural network composed of Izhikevich neurons, which is learned using STDP. Finally, fuzzy C-means clustering (FCM) is used to estimate heart rate, and the network achieves high precision and low power consumption.

Corradi et al. [36] used SNN to classify ECG heartbeat based on neuromorphic hardware. The idea is to compress the ECG signal directly into a four-channel spiking signal, and then input it into the recurrent spiking neural network for analysis and processing. Finally, its output is used as the input of the LIF classifier, and the category corresponding to the most active LIF neurons is the result of classification. The LIF classifier is first trained by a standard support vector machine, and then the numerical results are converted into the weight vector of LIF neurons. The results achieved over 95% accuracy across the 18 arrhythmias provided by the MIT-BIH dataset and supported hardware devices.

Rana et al. [37] proposed a binarized SNN to classify ECG signals. Binarized SNN means that the synaptic weight in the spiking neural network is set to 1 or -1 to further reduce the time and cost required for training. The experimental accuracy is lower than that of traditional DNNs, but the power is further reduced than that of conventional SNNs, and the accuracy of experimental results in this paper is 3% higher than that of traditional SNNS.

## 4 The Analysis of Electromyography (EMG)

Electromyography is mainly used to detect peripheral nerves and muscles. By monitoring the state of muscles, it can meet the specific needs of humans, such as judging fatigue driving of car drivers, and helping disabled people complete daily life. Surface EMG and other physiological signals can be combined to better meet these needs.

### 4.1. Pattern Recognition

Peng et al. [38] used the NeuCube framework to realized the recognition of six gestures, and combined the time-domain features with the NeuCube framework to achieve 95.33% accuracy. However, in the absence of manual features, the accuracy of classification of raw surface EMG data is only 68.7%. Behrenbeck et al. [39] also realized the classification of four kinds of surface EMG signals based on the NeuCube model, with a correct rate of 85% and a verification accuracy of 84.8%.

Donati et al. [40] used SNNs to classify EMG signals. Firstly, the EMG information corresponding to scissors, stone and cloth was collected from ten healthy subjects, after that, the original EMG signal was encoded as a spiking sequence, and then the spiking sequence was input into the dynamic neuromorphic asynchronous processing chip of SNNs. Finally, three methods were adopted for classification: Support vector machine SVM, Logistic regression method and spiking classification, the test results show that accuracy of the first two reached 84% and 81% respectively, while the accuracy of the spiking classificationwas 74%, but its power consumption is much lower.

Tieck et al. [41] used surface EMG to classify bent fingers, that is, to detect which finger is bent, and then use the classification results to achieve mechanical finger bending. In the classification process, the original signal is converted into spiking signal by pre-processing, and then supervised off-line training is carried out. Finally, the active finger is recognized online and human-computer interaction is realized. However, it may be difficult for most people to bend a finger alone, for example, the ring finger may have some degrees of bending when bending the middle finger, so the article suggested to further focus on the degree of bending.

Ma et al. [42] developed an SRNN system for online electromyography classification, which achieved gesture classification of electromyography. The main work is divided into three steps: spiking encoding, feature extraction, and classification. The experimental classification performance of the rosambo electromyography dataset has reached over 85%. The classification performance for the Ninapro DB2 database reached 55%. Although the classification accuracy of the Ninapro DB2 database is low, the network has the advantages of low latency and low power consumption, providing ideas for developing compact embedded and ultra-low power electronic systems that can classify EMG time data in real-time.

*4.2. Fatigue recognition*

Xu et al. [43] applied EMG to fatigue classification of crane operators. By measuring the surface EMG of the four points of the gastrocnemius and brachioradialis muscle, the EMG of the driver in the normal state, the critical state and the fatigue state can be determined to classify whether the staff is fatigue driving.

**5 Magnetic resonance imaging (MRI) analysis**

MRI imaging is one of the main research areas of medical image analysis. In recent years, deep learning performed well in the analysis of MRI, for example, Lv et al. [44] achieved early diagnosis of Alzheimer's disease based on enhanced AlexNet. Since SNNs consume less computing energy and are more bioexplainable, using SNNs for MRI research is of great significance. At present, research in this area includes the classification of functional MRI images by fMRI, the prediction and classification of Alzheimer's disease, the classification of brain tumors and the segmentation of brain tumor regions.

*5.1 Classification of fMRI*

Doborjeh et al. [45] classified the online functional NMR images based on NeuCube evolutionary spiking neural network. By letting the subjects read positive and negative sentences, using the visualization of the NeuCube model, the NMR classification is performed according to the different states and active regions of the network under different types of sentences, and the experimental results reach an average accuracy of 90%, which is better than traditional machine learning. In addition, the classification accuracy of negative sentences is 100%, and the accuracy of positive sentences is 80%, indicating that negative sentences are easier to distinguish, and it can be observed in the NeuCube model that the left brain is more active than the right brain, a finding that can help us further understand the functions of brain.

Kasabov et al. [46] further studied two classifications of fMRI with NeuCube. The first is the classification of sentences, that is, the classification of positive and negative sentences mentioned above, and the second is the classification of reading sentences and observing pictures, of which the classification of pictures reaches 80%, which further illustrates the different activities and functions of the brain under different conditions.

*5.2 Alzheimer's disease*

Doborjeh et al. [47] constructed a three-dimensional personalized spiking neural network (PSNN) based on longitudinal (cross-time) NMR image data using unsupervised STDP rules, and conducted prediction and classification based on the dataset. The experimental dataset consisted of 175 samples, all of which were collected three times over a period of six years without dementia. The preprocessed MRI data is first converted into a pulse train, and then mapped to a three-dimensional PSNN network of brain-like structures, trained using unsupervised STDP rules, and finally supervised learning using Dynamic Evolution SNN (DeSNN). The experimental results showed that the prediction of cognitive decline (mild cognitive impairment and Alzheimer's disease) two years in advance was 90% accurate.

Alzheimer's disease (AD) is a type of dementia that causes mental degeneration. Mild cognitive impairment (MCI) is a condition that falls somewhere between AD and healthy. Turkson et al. [48] designed a neural network using pulsed depth convolutional neural network to classify NMR images of Alzheimer's disease. MRI images included healthy controls, MCI patients, and AD patients. The classification task is transformed into three binary tasks (AD vs. NC, AD vs. MCI, and NC vs. MCI). The method includes three main steps: preprocessing, feature extraction and classification. Unsupervised convolutional SNN is used for AD feature extraction, and supervised pulse deep convolutional neural network CNN is used for classification. The experimental results reached 90.15%, 87.30% and 83.90% accuracy, respectively, which proved that the effect of feature extraction using SNN was superior to traditional machine learning.

*5.3 Brain tumors*

Baladhandapani et al. [49] used spiking neural networks to classify brain tumors. Firstly, a multidimensional symbiotic matrix is used for feature extraction, and then a genetic algorithm is used to train a spiking neural network classifier. It is verified on the Harvard benchmark dataset and the real-time dataset of the hospital, better results than the traditional feedforward backpropagation network is obtained. Experimental results show that the accuracy of the feedforward backpropagation classifier is 82.85%, and the best accuracy of the SNN classifier is 97.4%. However, the characteristics of the multidimensional symbiotic matrix itself may leads to misclassification between large benign tumors and malignant tumors, so further improvement is needed.

Ahmadi et al. [50] propose an MRI-based method for brain tumor segmentation: QAIS-DSNN. The experiment is divided into two parts, for the BraTS2018 dataset, preprocessing is first used to reduce noise, and quantum matching filtering technology (QMFT) is used as a local search operator to enhance the initial image. The second part is the segmentation stage using the combination of deep spiking neural networks (DSNNs) and quantum artificial immune systems (QAIS): first a convolutional and pooling layer that integrates CRF is employed for training, then it enters the SoftMax layer, uses the quantum artificial immune system (QAIS) to optimize segmentation and detection, and finally outputs tumor results. With 75% training set and 25% test set of the BraTS2018 dataset, the final accuracy was 98.21%, which has better performance advantages than similar methods, and shortens the calculation time with the execution time of only 2.58 seconds.

**6. Other applications**

*6.1. Obstacle avoidance for the blind*

Ge et al. [51] used the Neucube model to classify two types of samples, dynamic obstacles and static obstacles, respectively, and used grid search for parameter optimization of the Neucube model. The experiments showed that the overall accuracy, true positive rate and true negative rate of 40 samples from the static database were 90.5%, 88% and 93%, respectively; the detection of dynamic obstacles was more difficult and therefore the accuracy decreases to 81%, 86% and 76%, respectively, but the overall improvement over traditional machine learning is significant.

*6.2. Detection of abnormal gastric electric rhythm*

Breen et al. [52] used the NeuCube model to classify three common abnormal gastric electric rhythms and normal conditions based on electrogastrograms of the stomach (EGG). Three coding methods were used for comparison and the EGG signals were converted into pulse sequences for input into the NeuCube model. The results showed that the moving window (MW) method gave the most consistent and accurate classification results. The structure of the NeuCube model uses the spatial location of the electrodes placed in the body as the input nodes, and the computational nodes were spatially separated from the input nodes, with the final model containing 851 input nodes and 1000 computational nodes, achieving a maximum accuracy of 100% in all four classifications.

*6.3. Detection of respiratory abnormalities in preterm infants*

Paul et al. [53] used a CNN to SNN conversion method to classify respiratory abnormalities in preterm infants. The results showed that the accuracy of the 1DCNN classification reached 97.15%, SNN reduced power consumption by 18 times with slightly reduced accuracy of 93.33%, and reduced power consumption by 4 times without changing accuracy.

*6.4. Breast cancer (BrC) detection*

According to the official website of WHO's International Agency for Research on Cancer (IARC), breast cancer accounted for 11.7% of all new cancer cases worldwide in 2020, making it the most common cancer worldwide. Breast tumors are broadly classified into benign ones and malignant ones, and early detection will facilitate later treatment and improve patient survival. In recent years, there has been an increasing number of studies on classifications of breast tumors, and the main research methods include CNNs and SNNs [54-56].

Fu et al. [57] constructed a network capable of breast cancer identification based on SNN. The network uses the LIF neuron model and is divided into four parts: preprocessing, input layer, reservoir layer, and readout layer. The preprocessing uses the spiking convolutional neural network for feature extraction - the 2D feature map obtained by the spiking convolutional neural network is first used to calculate the mask map, and then the dot product of the original image and the mask image is calculated to generate the significant feature map. The input layer uses two types of temporal coding - linear temporal coding and entropy-based temporal coding - to input the pulse sequence into the reservoir layer. The final readout neurons are trained using the ReSuMe algorithm [58, 59]. The accuracy of the model is higher than 95% in the BreastMNIST database [60], and mini-MIAS database [61], which improves the classification accuracy.

**7. Conclusion**

In summary, spiking neural network has broad prospects in the field of medical health. In this paper, bases on medical data such as EEG, ECG, EMG, MRI, etc., the application and research results of spiking neural network are presented. Compared with traditional neural network, these research all achieved low power consumption with basically no loss of accuracy.

Generally, most of these models use LIF neurons and Izhikevich neuron models, and the learning rules are mainly based on STDP, in addition to gradient descent algorithm and ReSuMe algorithm. The deep learning networks involved are mainly recurrent neural networks dominated by NueCube and impulsive convolutional neural networks that convert CNNs into SNNs. Possible future directions of spiking neural network include:

1. By analyzing previous articles, we find that most of the research focused on offline learning of spiking neural networks. Online learning can respond to new data environment changes in a timely manner and learn faster. At the same time, as more and more data are available, online learning can be trained continuously using more data and the model will be updated continuously to solve problems in practical applications and better realize human-computer interaction, so it is important to expand the research on online learning.

2. The NeuCube model not only achieves good classification results, but also visualizes the model to help us understand and analyze brain functions. Inspired by this, spiking neural networks, as the intersection of computer technology and biomedicine, may achieve unexpected experimental results when conducting model design by combining traditional machine learning-related techniques with biomedical knowledge.

**Acknowledgements**

This research is supported by National Natural Science Foundation of China (No. 12171482) and State Key Laboratory of Petroleum Resources and Prospecting, China University of Petroleum (No. PRP/DX-2307).